\documentclass[a4paper,twoside]{article}

\usepackage{epsfig}
\usepackage{subcaption}
\usepackage{calc}
\usepackage{amssymb}
\usepackage{amstext}
\usepackage{amsmath}
\usepackage{amsthm}
\usepackage{multicol}
\usepackage{pslatex}
\usepackage{apalike}
\usepackage{algorithm2e}
\usepackage[bottom]{footmisc}

\usepackage{cite}
\usepackage{amsfonts}
\usepackage{algorithmic}
\usepackage{graphicx}
\usepackage{textcomp}
\usepackage{xcolor}

\usepackage{hyperref}
\usepackage[nolist]{acronym}
\usepackage{bm}
\usepackage{siunitx}
\usepackage{makecell}

\hypersetup{
    pdftitle={Towards UAV-USV Collaboration in Harsh Maritime Conditions Including Large Waves},
    pdfpagemode=FullScreen,
    pdfauthor={Filip Novak, Tomas Baca, Ondrej Prochazka, Martin Saska},
}

\usepackage{tikz}
\usepackage{pgfplots}
\pgfplotsset{compat=1.14}

\usetikzlibrary{quotes, angles, backgrounds, arrows, automata, shapes, positioning, calc, through, spy, decorations.pathreplacing, decorations.markings, arrows.meta, automata, petri, intersections, pgfplots.fillbetween}

\usetikzlibrary{backgrounds,arrows,automata,shapes,positioning,calc,through,spy}
\pgfdeclarelayer{background}
\pgfdeclarelayer{foreground}
\pgfsetlayers{background, main, foreground}

\tikzset{
  state/.style={
    rectangle,
    draw=black, very thick,
    minimum height=1.0em,
    text centered,
  },
  smallstate/.style={
    rectangle,
    draw=black, very thick,
    minimum height=0.2em,
    text centered,
  },
  final_state/.style={
    rectangle,
    rounded corners,
    draw=black, very thick,
    minimum height=2em,
    text centered,
  },
  initial_state/.style={
    rectangle,
    double=white,
    double distance=1pt,
    inner sep=2pt,
    draw=black, very thick,
    minimum height=2em,
    text centered,
  },
  point/.style={
    circle,
    inner sep=0pt,
    minimum size=3pt,
    fill=red
  },
  adder/.style={
    circle,
    inner sep=2pt,
    minimum size=0.3in,
    draw=black, very thick,
    text centered
  },
  state_gray/.style={
    rectangle,
    draw=black, very thick,
    fill=gray!40,
    minimum height=1.0em,
    text centered,
    inner sep=0,
  },
  state_white/.style={
    rectangle,
    draw=black, very thick,
    fill=white,
    minimum height=1.0em,
    text centered,
    text=black,
    inner sep=0,
  },
  state_green/.style={
    rectangle,
    draw=black, very thick,
    fill=green!50,
    minimum height=1.0em,
    text centered,
    text=black,
    inner sep=0,
  },
  state_red/.style={
    rectangle,
    draw=black, very thick,
    fill=red!70,
    minimum height=1.0em,
    text centered,
    text=black,
    inner sep=0,
  },
  state_blue/.style={
    rectangle,
    draw=black, very thick,
    fill=blue!40,
    minimum height=1.0em,
    text centered,
    text=black,
    inner sep=0,
  },
  final_state/.style={
    rectangle,
    rounded corners,
    draw=black, very thick,
    minimum height=2em,
    text centered,
  },
  initial_state/.style={
    rectangle,
    double=white,
    double distance=1pt,
    inner sep=2pt,
    draw=black, very thick,
    minimum height=2em,
    text centered,
  },
  point/.style={
    circle,
    inner sep=0pt,
    minimum size=3pt,
    fill=red
  },
}

\tikzset{new spy style/.style={spy scope={%
  magnification=5,
  size=1.25cm,
  connect spies,
  every spy on node/.style={
    rectangle,
    draw,
  },
  every spy in node/.style={
    draw,
    rectangle,
    fill=white
  }
  }
  }
}

\usepackage{SCITEPRESS}     

\renewcommand{\vec}[1]{\bm{#1}}
\newcommand{\mat}[1]{\mathbf{#1}}
\newcommand{\sota}{SOTA}

\newcommand{\reffig}[1]{Fig.~\ref{#1}}

\newcommand{\refsec}[1]{Sec.~\ref{#1}}

\newcommand{\reftab}[1]{Table~\ref{#1}}
\newcommand{\refeq}[1]{\eqref{#1}}

\newcommand{\worldframe}[0]{\mathcal{W}}
\newcommand{\worldframeaxis}[1]{\mathbf{w}_{#1}}
\newcommand{\bodyframe}[1]{\mathcal{B}_{#1}}
\newcommand{\bodyframeaxis}[2]{\mathbf{b}_{#1,#2}}

\newcommand{\vesselparallelframe}[0]{\mathcal{VP}}
\newcommand{\vesselparallelframeaxis}[1]{\mathbf{vp}_{#1}}

\newcommand{\PREPRINTYEAR}{2024}
\newcommand{\PUBLISHEDIN}{International Conference on Informatics in Control, Automation and Robotics (ICINCO 2024)}
\newcommand{\DOI}{10.5220/0012910000003822} 

\usepackage[placement=top,vshift=-7]{background}
\SetBgScale{1.0}
\SetBgContents{ICINCO 2024. PREPRINT VERSION - DO NOT DISTRIBUTE. \href{https://doi.org/\DOI}{DOI \DOI}}
\SetBgColor{black}
\SetBgAngle{0}
\SetBgOpacity{1.0}

\begin{document}

\thispagestyle{empty}
\onecolumn
{
  \topskip0pt
  \vspace*{\fill}
  \centering
  \LARGE{%
    \copyright{} \PREPRINTYEAR~\PUBLISHEDIN\\\vspace{1cm}
    Personal use of this material is permitted.
    Permission from \PUBLISHEDIN\footnote{\url{https://icinco.scitevents.org}}~must be obtained for all other uses, in any current or future media, including reprinting or republishing this material for advertising or promotional purposes, creating new collective works, for resale or redistribution to servers or lists, or reuse of any copyrighted component of this work in other works.
    \vspace{1cm}\newline
    \large DOI: \href{https://doi.org/\DOI}{\DOI}
    \vspace*{\fill}
    }
    \vspace*{\fill}
}
\NoBgThispage
\twocolumn          	
\BgThispage
\setcounter{footnote}{0}

\title{Towards UAV-USV Collaboration in Harsh Maritime Conditions Including Large Waves}

\author{
\authorname{
    Filip Nov\'{a}k\orcidAuthor{0000-0003-3826-5904}, 
    Tom\'{a}\v{s} B\'{a}\v{c}a\orcidAuthor{0000-0001-9649-8277},
    Ond\v{r}ej Proch\'{a}zka\orcidAuthor{0009-0009-2224-750X}, and
    Martin Saska\orcidAuthor{0000-0001-7106-3816} 
}
\affiliation{Department of Cybernetics, Faculty of Electrical Engineering, Czech Technical University in Prague} \email{filip.novak@fel.cvut.cz}
}


\abstract{
This paper introduces a system designed for tight collaboration between Unmanned Aerial Vehicles (UAVs) and Unmanned Surface Vehicles (USVs) in harsh maritime conditions characterized by large waves.
This onboard UAV system aims to enhance collaboration with USVs for following and landing tasks under such challenging conditions.
The main contribution of our system is the novel mathematical USV model, describing the movement of the USV in 6 degrees of freedom on a wavy water surface, which is used to estimate and predict USV states.
The estimator fuses data from multiple global and onboard sensors, ensuring accurate USV state estimation.
The predictor computes future USV states using the novel mathematical USV model and the last estimated states.
The estimated and predicted USV states are forwarded into a trajectory planner that generates a UAV trajectory for following the USV or landing on its deck, even in harsh environmental conditions.
The proposed approach was verified in numerous simulations and deployed to the real world, where the UAV was able to follow the USV and land on its deck repeatedly.
}

\onecolumn \maketitle \normalsize \setcounter{footnote}{0} \vfill

\section{\uppercase{Introduction}}
The \acp{UAV} have already proven their efficiency in numerous marine applications.
The \acp{UAV} are helpful in search and rescue operations \cite{usv_uav_hurricane_wilma}, monitoring marine animals \cite{marine_mammals_monitoring}, monitoring water quality \cite{water_quality_monitoring}, or cleaning oceans from garbage and oil spills \cite{water_pollution}.
However, the \acp{UAV} are limited by short battery life, which reduces their operational time.
This limitation is problematic as marine tasks are often executed at long distances from offshore base stations.
Therefore, the \acp{UAV} often collaborate with the \acp{USV} \cite{water_pollution, usv_uav_hurricane_wilma2}, which can compensate for \acp{UAV} short battery life by power umbilical tether \cite{talke2018CatenaryTetherShape} or providing docking spot for battery recharging \cite{usv_uav_docking}. 

In order to provide power supply via an umbilical tether from the \ac{USV} to the \ac{UAV}, the \ac{UAV} has to precisely follow the \ac{USV} at the specified distance based on the tether length \cite{talke2018CatenaryTetherShape}, which requires estimating and predicting \ac{USV} movement even in harsh conditions such as rough waters with large waves.
Similarly, estimating and predicting \ac{USV} movement is crucial for the \ac{UAV} during landing on the \ac{USV} docking spot \cite{uav_usv_landing_parakh}.
The oscillating and tilting \ac{USV} on a wavy water surface can significantly damage the landing \ac{UAV} or even cause \ac{UAV} to fall into the water.

\begin{figure}[!t]
    \centering
    \includegraphics[width=\linewidth]{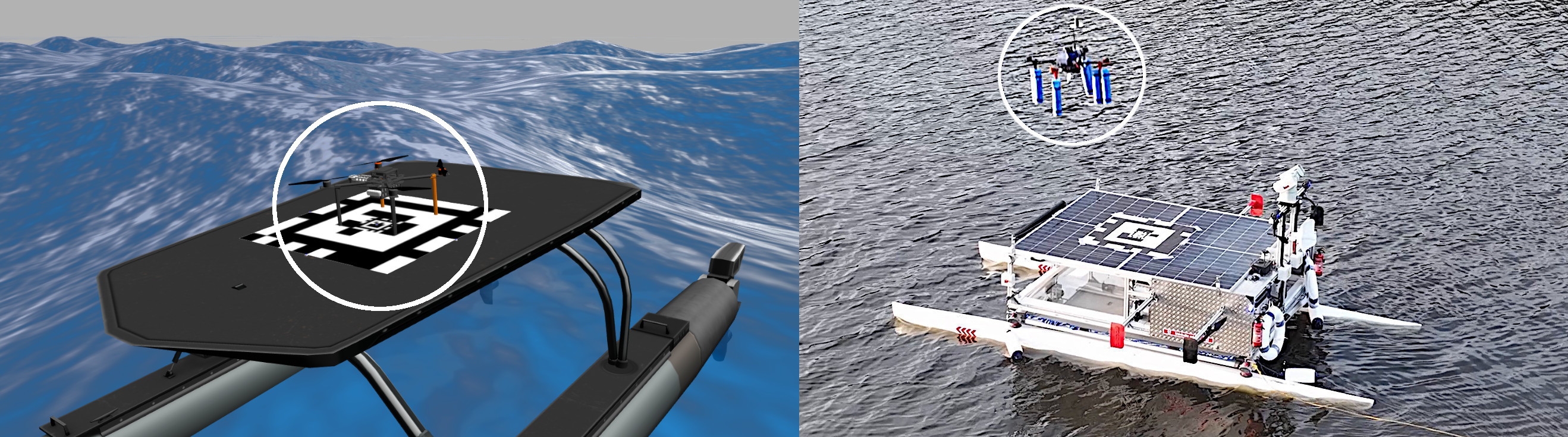}
    \caption{The tight collaboration between UAVs (marked with white circles) and USVs during the landing and following tasks using our system presented in this paper.}
    \label{fig:intro}
\end{figure}

The approach proposed in this paper allows tight collaboration between \acp{UAV} and \acp{USV} (see \reffig{fig:intro}), such as following and landing tasks on rough water surfaces.  
The key component of the presented system is the \ac{USV} state estimator, which runs onboard the \ac{UAV} and fuses data from the onboard sensors of both robots.
Using a novel mathematical \ac{USV} model containing wave dynamics, the estimated states are used to predict future \ac{USV} movement on the wavy water surface.
The estimated and predicted \ac{USV} states enable precise \ac{UAV} trajectory planning for following and landing.
The proposed approach was verified in numerous simulations and real-world experiments in both use cases: following the \ac{USV} and landing on the \ac{USV} docking spot.

The main paper contributions are summarized in the following points:
\begin{itemize}
    \item We propose a novel linear \ac{USV} mathematical model containing wave dynamics that enables accurate estimation and prediction of the \ac{USV} movement on wavy water surface in 6 \acp{DOF}.
    \item We introduce a novel onboard \ac{UAV} system to tight collaboration between \acp{UAV} and \acp{USV} in following and landing tasks in harsh maritime conditions, including large waves.
\end{itemize}


\section{\uppercase{Related Works}}
The works \cite{MENG2019474, lee2019} propose solutions for the following and landing of fixed-wing \acp{UAV} on marine vessels.
The method presented in \cite{MENG2019474} uses the auto-regressive model to predict the landing pad position at the touchdown moment.
The landing pad is equipped with infrared targets that are detected onboard \ac{UAV} to measure the relative position of the landing pad. 
The approach \cite{lee2019} relies on a sliding-mode control scheme to guide the UAV towards the landing point by following a desired reference trajectory.
The position of the landing point at touchdown time is predicted to be used in the final phase of the landing.
However, neither of these methods has been verified through real-world experiments, and the harsh environment is not considered in simulations. 

An internal-model-based approach for \ac{VTOL} \acp{UAV} is introduced in \cite{MARCONI200221}.
This approach is designed for autonomous landing on a vertically oscillating deck. 
The oscillations of the deck are modeled as a sum of sinusoidal functions.
However, the oscillations are considered only in heave motion.
The approach does not consider the pitch and roll motions of the landing deck.
Furthermore, the approach is not deployed in the real world, and its validation is limited to simulations.

A visual-based autonomous landing of \ac{UAV} on a moving \ac{USV} is presented in \cite{keller2022}.
The method \cite{keller2022} relies only on the position of the landing platform. 
However, it's important to note that this method does not account for the roll and pitch motions during landing, which can pose a risk, particularly in rough water conditions where the rolling and pitching of the landing platform may potentially damage the \ac{UAV}.
Although the method \cite{keller2022} is verified in real-world experiments, the method is not designed for harsh environments.
A similar vision-based approach is presented in \cite{Venugopalan2012}.
This approach also considers only \ac{USV} position as \cite{keller2022} and is deployed in the real world.
Neither methods \cite{keller2022, Venugopalan2012} analyze waves influencing the \ac{USV} motions, which does not make them prepared for harsh conditions.

The method for following and landing verified in the real world is presented in \cite{Xu2020_vision}.
In this method, the \ac{UAV} uses a camera to detect a tag placed on the landing platform, which is placed onboard \ac{USV}.
The detected tag is then used to estimate the relative position from the \ac{UAV} to the \ac{USV}.
A similar estimation method is also employed in \cite{yang2021}.
Both methods are verified by conducting real-world experiments.
However, harsh conditions, such as large waves, as well as pitch and roll motions of the \ac{USV}, are not considered in these methods.

The complex system estimating \ac{USV} motion in 5 \acp{DOF} is proposed in \cite{Abujoub2018_landing}.
The estimation method differs from previous vision-based approaches by utilizing \ac{LiDAR} to measure the position and orientation of the \ac{USV}.
Additionally, the future roll and pitch motions are predicted as the sum of harmonic functions representing the wave motions.
The parameters of these harmonic functions are derived from \ac{FFT} of estimated roll and pitch angles.
However, this method does not consider wave motion in the heave state of the \ac{USV}, which could potentially lead to unsafe landing or following, especially in rough conditions.
Furthermore, the method is solely tested in simulations and has not been validated in real-world experiments.

The above-mentioned methods \cite{keller2022, Venugopalan2012, Xu2020_vision, yang2021, Abujoub2018_landing} rely solely on vision methods to estimate \ac{USV} states. 
However, a significant limitation of these methods arises when the \ac{USV} falls outside the \ac{FOV} of the \ac{UAV} sensors, rendering it impossible to estimate the position and orientation of the \ac{USV}.
In such cases, the \ac{UAV} must actively search for the \ac{USV}, leading to increased energy consumption and limiting the time dedicated to the \ac{UAV} mission.
An alternative approach, as presented in \cite{zhang2021}, combines \ac{GNSS} sensors placed on the \ac{USV} with onboard \ac{UAV} vision-based systems to estimate the position of the \ac{USV}. However, it's important to note that neither the orientation of the \ac{USV} nor wave motions are accounted for in the estimation process. Consequently, the method presented in \cite{zhang2021} is deemed unsuitable for harsh conditions.

The method proposing \ac{USV} state estimation in full 6 \acp{DOF} is presented in \cite{polvara_6dof}.
The \ac{UAV} detects the tag placed on \ac{USV} board in camera images and utilizes it for the estimation.
Subsequently, the method controls the \ac{UAV} landing on the \ac{USV} board.
However, the waves are not considered for any \ac{USV} state, making the method unsuitable for harsh conditions that are considered in this paper.
The method for landing \ac{UAV} on \ac{USV} in harsh conditions is presented in \cite{uav_usv_landing_parakh}.
The method considers waves in the \ac{USV} motion and predicts future \ac{USV} states to perform a safe landing. 
However, the assumption made in \cite{uav_usv_landing_parakh} that the \ac{USV} is not moving horizontally on the water surface renders the method unsuitable for the task of \ac{UAV} following.
Additionally, both methods \cite{polvara_6dof, uav_usv_landing_parakh} are purely vision-based, causing aforementioned issues if the \ac{USV} is not in the \ac{FOV} of the \ac{UAV} onboard sensors. 

In comparison with the above-mentioned methods, our system integrates both \ac{USV} onboard sensors and \ac{UAV} onboard sensors to accurately estimate the \ac{USV} motion in full 6 \acp{DOF}, while factoring in wave dynamics at every state.
Additionally, our system predicts the future \ac{USV} states in 6 \acp{DOF}, incorporating wave motions. 
Therefore, our system enables tight \ac{UAV}-\ac{USV} collaboration in extreme environmental conditions at different relative \ac{UAV}-\ac{USV} distances constrained by the communication range.
The estimated and predicted \ac{USV} states serve as inputs to the \ac{UAV} trajectory planner, based on the \ac{MPC} method, enabling the \ac{UAV} to follow the \ac{USV} and land on it even in harsh environmental conditions.

\section{\uppercase{Mathematical USV model}}
\label{sec:mathematical_model}
\begin{figure}[!tb]
    \centering
    \includegraphics[width=0.75\linewidth]{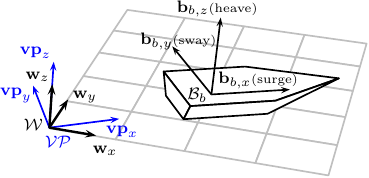}
    \caption{The depiction of the world frame $\worldframe = \{\worldframeaxis{x}, \worldframeaxis{y}, \worldframeaxis{z}\}$, Vessel parallel coordinate system $\vesselparallelframe=\{ \vesselparallelframeaxis{x},\vesselparallelframeaxis{y},\vesselparallelframeaxis{z}\}$, and USV body-fixed coordinate frame $\bodyframe{b} = \{\bodyframeaxis{b}{x}, \bodyframeaxis{b}{y}, \bodyframeaxis{b}{z}\}$.}
    \label{fig:coordinate_frames}
\end{figure}

We identify accurate state estimation and prediction of \ac{USV} states as crucial features for tight \ac{UAV}-\ac{USV} collaboration, enabling the \ac{UAV} to follow the \ac{USV} and land on its deck even in harsh conditions, including large waves.
In this section, we present a novel linear \ac{USV} model containing wave dynamics in order to fuse data from multiple sensors, thereby increasing estimation and prediction accuracy.
We model the \ac{USV} in 6 \acp{DOF}, consisting of 3D translation (surge~$x$, sway~$y$, and heave~$z$) and 3D rotation in terms of intrinsic Euler angles (roll $\phi$, pitch $\theta$, and yaw $\psi$), as illustrated in \reffig{fig:coordinate_frames}.

In order to analyze the \ac{USV} motion, three coordinate frames are presented: the world coordinate frame $\worldframe = \{\worldframeaxis{x}, \worldframeaxis{y}, \worldframeaxis{z}\}$, the body-fixed coordinate frame $\bodyframe{b} = \{\bodyframeaxis{b}{x}, \bodyframeaxis{b}{y}, \bodyframeaxis{b}{z}\}$, and the Vessel parallel coordinate frame $\vesselparallelframe=\{ \vesselparallelframeaxis{x},\vesselparallelframeaxis{y},\vesselparallelframeaxis{z}\}$ as shown in \reffig{fig:coordinate_frames}.
The position $\vec{p}_L = (x_L,y_L,z_L)^\intercal$ and rotation $\vec{\Theta}_L=(\phi_L,\theta_L,\psi_L)^\intercal$ of the \ac{USV} are expressed in the Vessel parallel coordinate frame $\vesselparallelframe$, which is parallel to the body-fixed coordinate frame $\bodyframe{b}$ and placed at the origin of the world coordinate frame $\worldframe$.
The \ac{USV} state, expressed in Vessel parallel coordinate frame $\vec{\eta}_L = (\vec{p}^\intercal_L,\vec{\Theta}^\intercal_L)^\intercal$, is transformed to the world coordinate frame as $\vec{\eta} = (\vec{p}^\intercal,\vec{\Theta}^\intercal)^\intercal$ as
\begin{align}
    \vec{\eta} = \mat{J}_\psi(\psi) \vec{\eta}_L,
\end{align}
where transformation matrix $\mat{J}_\psi(\psi)$ is defined as
\begin{align}
    \mat{J}_{\psi}(\psi) &= \begin{pmatrix}
     \mat{R}_{\psi} & \mat{O}_{3\times3}\\
     \mat{O}_{3\times3} & \mat{I}_{3\times3}
    \end{pmatrix},\\
    \mat{R}_\psi &= \begin{pmatrix}
        \cos \psi & -\sin \psi & 0\\
        \sin \psi & \cos \psi & 0\\
        0 & 0 & 1
    \end{pmatrix},
\end{align}
$\mat{O}_{3\times3}\in\mathbb{R}^{3\times3}$ is a zero matrix, and $\mat{I}_{3\times3}\in\mathbb{R}^{3\times3}$ denotes the identity matrix.
The Vessel parallel coordinate frame $\vesselparallelframe$ enables us to use an identity matrix as the transformation between velocity state vector $\vec{\nu}$ and the derivative of vector $\vec{\eta}_L$.
The linear velocity $\vec{v}=(u,v,w)^\intercal$ and angular velocity $\vec{\omega}=(p,q,r)^\intercal$ form the state vector $\vec{\nu} = (\vec{v}^\intercal,\vec{\omega}^\intercal)^\intercal$, expressed in the body-fixed coordinate frame $\bodyframe{b}$.

Our novel \ac{USV} model builds upon the \ac{USV} motion analysis presented in \cite{Fossen2011} and extends it by wave dynamics, resulting in a mathematical model that describes the \ac{USV} motion on a rough water surface.
First, the equations of \ac{USV} motions are
\begin{align}
    \vec{\dot{\eta}}_L &= \vec{\nu},\\
    \vec{\dot{\nu}} &= (\mat{M}_I+\mat{M}_A)^{-1}\left( -\mat{D}\vec{\nu} - \mat{G}\vec{\eta}_L \right),
\end{align}
where $\mat{M}_I\in\mathbb{R}^{6\times6}$ represents the inertia matrix, the $\mat{M}_A\in\mathbb{R}^{6\times6}$ is hydrodynamic added mass occurring due to the motion of the \ac{USV} through the fluid.
The matrix $\mat{D}\in\mathbb{R}^{6\times6}$ represents the linear damping and term $\mat{G}\in\mathbb{R}^{6\times6}$ denotes matrix of gravitational forces and torques, also called restoring forces.

We assume the waves as an oscillatory motion in each \ac{USV} state.
Therefore, we present one wave component as a 2 \ac{DOF} linear state-space model, defined by matrices $\mat{A}_\omega,~\mat{C}_\omega$ as
\begin{align}
    \begin{pmatrix}
     \dot{x}_{\omega_{1}}\\ \dot{x}_{\omega_{2}}
     \end{pmatrix} =& 
     \underbrace{
     \begin{pmatrix}
      0 & 1\\
      -\omega_{0}^2 & -2\lambda\omega_{0}
    \end{pmatrix}}_{\mat{A}_\omega}
    \begin{pmatrix}
    x_{\omega_{1}}\\ x_{\omega_{2}}
    \end{pmatrix}
    , \label{eq:linear_wave_model_x}
\end{align}
\begin{align}
    y_{\omega} &= \underbrace{\begin{pmatrix}
    0 & 1
    \end{pmatrix}}_{\mat{C}_\omega}
    \begin{pmatrix}
    x_{\omega_{1}}\\ x_{\omega_{2}}
    \end{pmatrix},
    \label{eq:linear_wave_model_y}
\end{align}
where $\vec{x}_\omega=(x_{\omega_{1}}, x_{\omega_{2}})^\intercal$ is a state of the wave component, $\lambda$ is a damping term, and $\omega_0$ denotes the frequency of the wave component.
To model wave motion with complex frequency spectra, we combined $N_{c}\in\mathbb{Z}^+$ wave components from equations \refeq{eq:linear_wave_model_x} and \refeq{eq:linear_wave_model_y}.
Each component is characterized by different parameters $\lambda$ and $\omega_0$
\begin{align}
    \vec{\dot{x}}_{\omega_{1}} &= \mat{A}_{\omega_{1}}\vec{x}_{\omega_{1}},\label{eq:linear_wave_system_first}\\
    y_{\omega_{1}} &= \mat{C}_{\omega_{1}}\vec{x}_{\omega_{1}},\\
    &~\vdots\nonumber\\
    \vec{\dot{x}}_{\omega_{N_{c}}} &= \mat{A}_{\omega_{N_{c}}}\vec{x}_{\omega_{N_{c}}},\\
    y_{\omega_{N_{c}}} &= \mat{C}_{\omega_{N_{c}}}\vec{x}_{\omega_{N_{c}}},\\
    y_{\text{wave}} &= y_{\omega_{1}} + \ldots + y_{\omega_{N_{c}}}.\label{eq:linear_wave_system_last}
\end{align}
The equations \refeq{eq:linear_wave_system_first}-\refeq{eq:linear_wave_system_last} can be expressed as
\begin{align}
    \vec{\dot{x}}_{\text{wave}} &= \mat{A}_{\text{wave}}\vec{x}_{\text{wave}},\label{eq:wave_statem_x}\\
    y_{\text{wave}} &= \mat{C}_{\text{wave}}\vec{x}_{\text{wave}},\label{eq:wave_statem_y}\\
    \vec{x}_{\text{wave}} &= (\vec{x}_{\omega_{1}}^\intercal,\ldots,\vec{x}_{\omega_{N_c}}^\intercal)^\intercal,\\
    \mat{A}_{\text{wave}} &= \text{diag}\{ \mat{A}_{\omega_{1}},\ldots,\mat{A}_{\omega_{N_c-1}} \},\\
    \mat{C}_{\text{wave}} &= \left( \mat{C}_{\omega_{1}}~\cdots~\mat{C}_{\omega_{N_c}} \right),
\end{align}
where $\text{diag}\{\cdot\}$ is a symbol for a block diagonal matrix created from the elements in the bracket.
The wave model \refeq{eq:wave_statem_x}--\refeq{eq:wave_statem_y} is integrated into each state of the \ac{USV} state vector $\vec{\nu}$.
Hence, we present the complex model of waves influencing \ac{USV} motion as
\begin{align}
    \vec{\dot{x}}_{\text{wave},\vec{\nu}} &= \mat{A}_{\text{wave},\vec{\nu}}\vec{x}_{\text{wave},\vec{\nu}},\label{eq:complex_wave_system_1}\\
    \vec{y}_{\text{wave},\vec{\nu}} &= \mat{C}_{\text{wave},\vec{\nu}}\vec{x}_{\text{wave},\vec{\nu}},
\end{align}
where $\mat{A}_{\text{wave},\vec{\nu}}$ and $\mat{C}_{\text{wave},\vec{\nu}}$ are block diagonal matrices
\begin{align}
    \mat{A}_{\text{wave},\vec{\nu}} &= \text{diag}\{ \mat{A}_{\text{wave}},\mat{A}_{\text{wave}},\mat{A}_{\text{wave}},\mat{A}_{\text{wave}},\nonumber\\&\mat{A}_{\text{wave}},\mat{A}_{\text{wave}} \}, \\
   \mat{C}_{\text{wave},\vec{\nu}} &= \begin{pmatrix}
        \mat{C}_{\text{wave}} & \mat{O}_{1\times 2N_c} & \ldots & \mat{O}_{1\times 2N_c}\\
        \mat{O}_{1\times 2N_c} & \ddots & \ddots & \vdots\\
        \vdots & \ddots & \mat{C}_{\text{wave}} & \mat{O}_{1\times 2N_c}\\
        \mat{O}_{1\times 2N_c} & \ldots & \mat{O}_{1\times 2N_c} & \mat{C}_{\text{wave}}
    \end{pmatrix},
\end{align}
and 
\begin{align}
    \vec{x}_{\text{wave},\vec{\nu}} = (\vec{x}_{\text{wave},u}^{\intercal},~\vec{x}_{\text{wave},v}^{\intercal},~\vec{x}_{\text{wave},w}^{\intercal}, \nonumber\\\vec{x}_{\text{wave},p}^{\intercal},~\vec{x}_{\text{wave},q}^{\intercal},~\vec{x}_{\text{wave},r}^{\intercal})^{\intercal}.
\end{align}

Finally, the novel 6 \ac{DOF} mathematical model of the \ac{USV} containing wave dynamics is
\begin{align}
    \vec{\dot{x}}_{\text{usv}} = \mat{A}_{\text{usv}}\vec{x}_{\text{usv}},\label{eq:linear_model_usv_wave}
\end{align}
where $\vec{x}_{\text{usv}} = (\vec{\eta}_{L}^{\intercal},~\vec{\nu}^{\intercal},~\vec{x}^{\intercal}_{\text{wave},\vec{\nu}})^{\intercal}$, 
\begin{align}
    \mat{A}_{\text{usv}} = \begin{pmatrix}
     \mat{O}_{6\times6} & \mat{I}_{6\times6} & \mat{O}_{6\times12N_c}\\
     -\mat{M}^{-1}\mat{G} & -\mat{M}^{-1}\mat{D} & \mat{C}_{\text{wave},\vec{\nu}}\\
     \mat{O}_{12N_c\times6} & \mat{O}_{12N_c\times6} & \mat{A}_{\text{wave},\vec{\nu}}
    \end{pmatrix},
\end{align}
and $\mat{M}=\mat{M}_I+\mat{M}_A$.

\begin{figure*}
  \centering
  \resizebox{1.0\textwidth}{!}{
   \pgfdeclarelayer{foreground}
\pgfsetlayers{background,main,foreground}

\tikzset{radiation/.style={{decorate,decoration={expanding waves,angle=90,segment length=4pt}}}}

\begin{tikzpicture}[->,>=stealth', node distance=3.0cm,scale=1.0, every node/.style={scale=1.0}]


  \node[state, shift = {(0.0, 0.0)}] (gps) {
      \begin{tabular}{c}
        \footnotesize GPS
      \end{tabular}
    };

    \node[state, below of = gps, shift = {(0.0, 2.0)}] (imu) {
      \begin{tabular}{c}
        \footnotesize IMU
      \end{tabular}
    };

  \node[state, right of = gps, shift = {(0, 0)}] (state_estimator) {
      \begin{tabular}{c}
        \footnotesize State \\
        \footnotesize estimator
      \end{tabular}
    };

  \node[state, right of = state_estimator, shift = {(0, 0)}] (state_predictor) {
      \begin{tabular}{c}
        \footnotesize State \\
        \footnotesize predictor
      \end{tabular}
    };

  \node[state, right of = state_predictor, shift = {(0, 0)}] (trajectory_planner) {
      \begin{tabular}{c}
        \footnotesize Trajectory\\
        \footnotesize planner
      \end{tabular}
    };

  \node[state, right of = trajectory_planner, shift = {(0, 0)}] (mrs_uav_system) {
      \begin{tabular}{c}
        \footnotesize MRS UAV\\
        \footnotesize system
      \end{tabular}
    };

  \node[state, right of = mrs_uav_system, shift = {(0, 0)}] (apriltag) {
      \begin{tabular}{c}
        \footnotesize AprilTag \\
        \footnotesize detector
      \end{tabular}
    };

    \node[state, below of = apriltag, shift = {(0.0, 1.5)}] (uvdar) {
      \begin{tabular}{c}
        \footnotesize UVDAR \\
        \footnotesize system
      \end{tabular}
    };



  \path[->] ($(gps.east) + (0.0, 0)$) edge [] node[above, midway, shift = {(-0.1, 0.0)}] {
      \begin{tabular}{c}
        \footnotesize 10~Hz
    \end{tabular}} ($(state_estimator.west) + (0.0, 0.00)$);

  \path[-] ($(imu.east) + (0.0, 0)$) edge [] node[above, midway, shift = {(0.0, 0.05)}] {
      \begin{tabular}{c}
    \end{tabular}} ($(imu.east)+(0.65,0)$) -- ($(imu.east)+(0.65,0)$) edge[-] ($(imu.east)+(0.65,0.875)$) -- ($(imu.east)+(0.65,0.875)$) edge[-] ($(imu.east)+(0.65,0.875)$) edge[->] ($(state_estimator.west)+(0.0,-0.125)$);

    \path[->] ($(state_estimator.east) + (0.0, 0)$) edge [] node[above, midway, shift = {(0.0, 0.0)}] {
      \begin{tabular}{c}
        \footnotesize 100~Hz
    \end{tabular}} ($(state_predictor.west) + (0.0, 0.00)$);

    \path[->] ($(state_estimator.east) + (0.0, -0.125)$)  edge[-] ($(state_estimator.east) + (0.5, -0.125)$) -- ($(state_estimator.east) + (0.5, -0.125)$) edge[-]($(state_estimator.east) + (0.5, -1.125)$) -- ($(state_estimator.east) + (0.5, -1.125)$) edge[-]
    ($(trajectory_planner.south) + (0, -0.55)$) --
    ($(trajectory_planner.south) + (0, -0.55)$)
    edge [] node[left, midway, shift = {(-0.25, -0.1)}] {
      \begin{tabular}{c}
        \footnotesize 100~Hz
    \end{tabular}} ($(trajectory_planner.south)$);

    \path[->] ($(state_predictor.east) + (0.0, 0)$) edge [] node[above, midway, shift = {(0.0, 0.0)}] {
      \begin{tabular}{c}
        \footnotesize 100~Hz
    \end{tabular}} ($(trajectory_planner.west) + (0.0, 0.00)$);

    \path[->] ($(trajectory_planner.east) + (0.0, 0)$) edge [] node[above, midway, shift = {(0.0, 0.0)}] {
      \begin{tabular}{c}
        \footnotesize 50~Hz
    \end{tabular}} ($(mrs_uav_system.west) + (0.0, 0.00)$);

    \path[-] ($(apriltag.west) + (0.0, 0)$) edge[-] ($(apriltag.west) + (-0.5, 0)$) -- ($(apriltag.west) + (-0.5, 0)$) edge[-] ($(apriltag.west) + (-0.5, -1.375)$) -- ($(apriltag.west) + (-0.5, -1.375)$) edge[-] ($(state_estimator.south |- uvdar.west)+(0.0625,0.125)$) -- ($(state_estimator.south |- uvdar.west)+(0.0625,0.125)$) edge[->] ($(state_estimator.south)+(0.0625,0)$);

    \path[-] ($(uvdar.west)$) edge [] node[below, midway, shift = {(0.1, 0.0)}] {
      \begin{tabular}{c}
        \footnotesize 30~Hz
    \end{tabular}} ($(state_estimator.south |- uvdar.west)+(-0.0625,0)$) -- ($(state_estimator.south |- uvdar.west)+(-0.0625,0)$) edge[->] ($(state_estimator.south)+(-0.0625,0)$);



  \begin{pgfonlayer}{background}
    \path (state_estimator.west |- state_estimator.north)+(-0.45,0.7) node (a) {};
    \path (trajectory_planner.south -| trajectory_planner.east)+(+0.25,-0.20) node (b) {};
    \path[fill=blue!5,rounded corners, draw=blue!70, densely dotted]
      (a) rectangle (b);
  \end{pgfonlayer}
  \node [rectangle, above of=state_predictor, shift={(0,0.35)}, node distance=1.7em] (autopilot) {\footnotesize UAV-USV collaboration};

  \begin{pgfonlayer}{background}
    \path (gps.west |- state_estimator.north)+(-0.25,0.7) node (a) {};
    \path (imu.south -| imu.east)+(+0.25,-0.30) node (b) {};
    \path[fill=black!3,rounded corners, draw=black!70, densely dotted]
      (a) rectangle (b);
  \end{pgfonlayer}
  \node [rectangle, above of=gps, shift={(0,0.2)}, node distance=1.7em] (autopilot) {\footnotesize USV plant};

  \begin{pgfonlayer}{background}
    \path (mrs_uav_system.west |- state_estimator.north)+(-0.25,0.7) node (a) {};
    \path (uvdar.south -| uvdar.east)+(+0.25,-0.10) node (b) {};
    \path[fill=black!3,rounded corners, draw=black!70, densely dotted]
      (a) rectangle (b);
  \end{pgfonlayer}
  \node [rectangle, above of=mrs_uav_system, shift={(1.35,0.3)}, node distance=1.7em] (autopilot) {\footnotesize UAV plant};


\end{tikzpicture}
   }
  \caption{
  The figure depicts a pipeline diagram of the entire system used for experimental verification in this paper.
  The \textit{State estimator} fuses data from USV onboard sensors (GPS and IMU) and UAV onboard sensors (AprilTag detector and UVDAR system).
  The estimated USV states are then sent to the \textit{State predictor}, which predicts future USV states.
  The \textit{Trajectory planner} uses the estimated and predicted USV states to generate a UAV trajectory, which is precisely tracked by the \textit{MRS UAV system} \cite{baca2021mrs}.
  }
  \label{fig:pipeline_diagram}
\end{figure*}
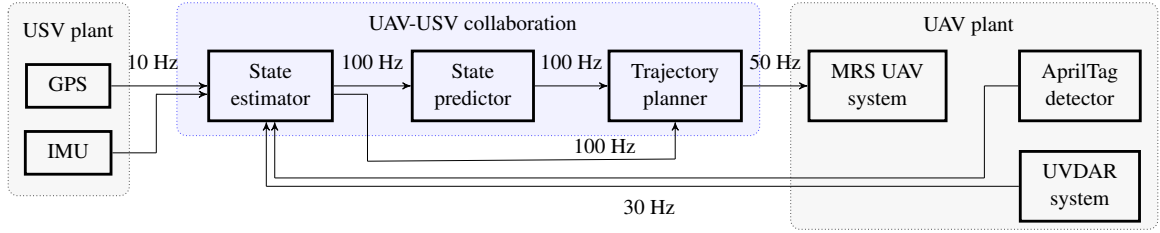

\section{\uppercase{UAV-USV collaboration in waves}}
The pipeline integrating the entire \ac{UAV}-\ac{USV} system is depicted in \reffig{fig:pipeline_diagram}.
The pipeline comprises three main areas -- \ac{USV} plant, \ac{UAV} plant, and the proposed approach for tight \ac{UAV}-\ac{USV} collaboration.
The \ac{USV} plan consists of a \ac{GPS} sensor and an \ac{IMU}, whose data are sent to our state estimator.
The \ac{UAV} plant includes onboard sensors -- AprilTag detector \cite{olson2011tags, wang2016iros, krogius2019iros},  and \ac{UVDAR} system \cite{walter_icra2020, uvdd1, uvdd2}, which are discussed later in this paper.
The \ac{UAV} is controlled by the \ac{MRS} \ac{UAV} system \cite{baca2021mrs}, enabling precise following of a trajectory planned by the proposed method.
The State estimator that fuses data from \ac{USV} and \ac{UAV} onboard sensors using a novel mathematical \ac{USV} model \refeq{eq:linear_model_usv_wave} to obtain an accurate estimate of \ac{USV} states moving on rough water surfaces.
The estimated states and novel mathematical model are then used in the State predictor to predict future \ac{USV} states.
The estimated and predicted \ac{USV} states are forwarded to the Trajectory planner, which generates a \ac{UAV} trajectory to follow the \ac{USV} and land on its deck.

\subsection{State Estimator and Predictor}
\label{sec:state_estimator_predictor}
We rely on the \ac{LKF} \cite{kalman1960new} as the state estimator, utilizing a discrete version $\mat{A}_{\text{usv},d}$ of our novel linear \ac{USV} model \refeq{eq:linear_model_usv_wave}.
The \ac{LKF} consists of two main steps: the prediction step and the correction step.
The prediction step uses the last estimated \ac{USV} state and propagates it through the mathematical model to obtain a new state
\begin{align}
    \vec{x}_{\text{usv}}(k+1) &= \mat{A}_{\text{usv},d}\vec{x}_{\text{usv}}(k),\\
    \mat{P}_{\text{usv}}(k+1) &= \mat{A}_{\text{usv},d}\mat{P}_{\text{usv}}(k)\mat{A}_{\text{usv},d}^{\intercal} + \mat{Q}_{\text{usv}},
\end{align}
where $k\in\mathbb{Z}^{+}$ is a time step, $\mat{P}_{\text{usv}}(k)\in\mathbb{R}^{12(1+N_c)\times12(1+N_c)}$ stands for covariance matrix of \ac{USV} state $\vec{x}_{\text{usv}}(k)$, and $\mat{Q}_{\text{usv}}\in\mathbb{R}^{12(1+N_c)\times12(1+N_c)}$ represents a system noise matrix.
The correction step incorporates incoming measurement $\vec{z}(k)$ to update the last estimated state
\begin{align}
    \vec{x}_{\text{usv}}(k) &= \vec{x}_{\text{usv}}(k) + \mat{G}(k)\left( \vec{z}(k) - \mat{C}\vec{x}_{\text{usv}}(k) \right),\\
    \mat{P}_{\text{usv}}(k) &= \mat{P}_{\text{usv}}(k) - \mat{G}(k)\mat{C}\mat{P}_{\text{usv}}(k),\\
    \mat{G}(k) &= \mat{P}_{\text{usv}}(k)\mat{C}^\intercal \left( \mat{C}\mat{P}_{\text{usv}}(k)\mat{C}^\intercal + \mat{R} \right)^{-1},
\end{align}
where $\mat{R}$ is measurement noise matrix, and matrix $\mat{C}$ represents mapping between state $\vec{x}_{\text{usv}}$ and measurement $\vec{z}$.
To predict future \ac{USV} states from the last estimated values $\vec{x}_{\text{usv}}(k)$ and $\mat{P}_{\text{usv}}(k)$, the prediction step of the \ac{LKF} is iteratively applied to obtain $N_{p}\in\mathbb{Z}^{+}$ number of predictions
\begin{align}
    \vec{\hat{x}}_{\text{usv}}(k_p+1) &= \mat{A}_{\text{usv},d}\vec{\hat{x}}_{\text{usv}}(k_p),\\
    \mat{\hat{P}}_{\text{usv}}(k_p+1) &= \mat{A}_{\text{usv},d}\mat{\hat{P}}_{\text{usv}}(k_p)\mat{A}_{\text{usv},d}^{\intercal} + \mat{Q}_{\text{usv}},
\end{align}
where $k_p=0,1,\ldots,N_p-1$, $\vec{\hat{x}}_{\text{usv}}(0)=\vec{x}_{\text{usv}}(k)$, and $\mat{\hat{P}}_{\text{usv}}(0)=\mat{P}_{\text{usv}}(k)$.

\subsection{Onboard USV Sensors}
\label{sec:usv_onboard_sensors}
\begin{figure}[!b]
    \centering
    \includegraphics[width=0.9\linewidth]{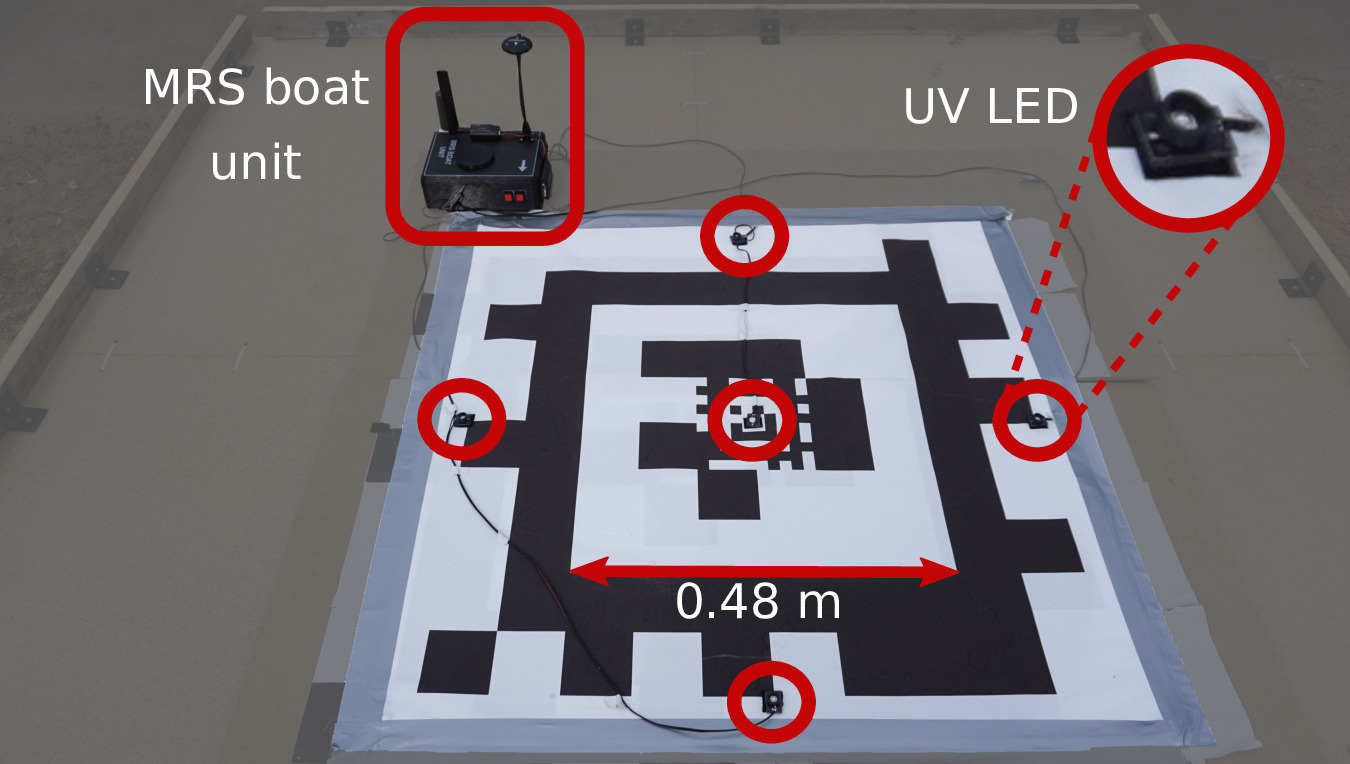}
    \caption{The USV board showing an AprilTag, UV LED (marked with red circles), and MRS boat unit, which contains GPS and IMU.}
    \label{fig:sensors}
\end{figure}

In practical scenarios, the \ac{UAV} often operates at distance where its onboard sensors cannot provide sufficient data for \ac{USV} state estimation. 
To address this issue, we assume a communication link between the \ac{UAV} and \ac{USV}, operating at least on \SI{10}{\hertz}.
The \ac{USV} sends sensor data to the \ac{UAV} to roughly estimate the \ac{USV} state.
Subsequently, the \ac{UAV} can fly to the proximity of the \ac{USV} to use onboard \ac{UAV} sensors, thereby increasing estimation and prediction precision.
The first onboard \ac{USV} sensor is a \ac{GPS} device providing global position information.
The second onboard \ac{USV} sensor is an \ac{IMU} that measures heading, angular velocity, and linear acceleration.
These \ac{USV} sensors are integrated within our \ac{MRS} boat unit, which is placed on the \ac{USV} board, as depicted in \reffig{fig:sensors}.

\subsection{Onboard UAV Sensors}
\label{sec:uav_onboard_sensors}
Establishing a fast and reliable communication link in the real world is challenging \cite{TRAN201967}.
To deal with drop-outs of communication links, the \ac{UAV} carries two onboard sensors that do not rely on a communication link.
These onboard \ac{UAV} sensors increase redundancy and enable usage of the system in various real-world conditions.
Moreover, the two onboard sensory modalities demonstrate the system's ability to fuse data from multiple \ac{UAV} and \ac{USV} sensors that can be chosen according to the desired application.

The first sensor is the AprilTag detector \cite{olson2011tags, wang2016iros, krogius2019iros}.
This vision-based system detects landmarks known as AprilTag in the camera image frame.
The used landmark placed on \ac{USV} board is shown in \reffig{fig:sensors}.
In this custom landmark layout, a smaller AprilTag is placed within the empty space of the larger AprilTag, enabling detection from various distances \cite{krogius2019iros}.
The measured data from the AprilTag detector includes the position and orientation of the detected landmark.
However, the AprilTag detector relies on sufficient lighting conditions to provide quality measurements due to its passive landmarks, which limits the system's usage, for example, in dark environments.

The second onboard \ac{UAV} sensor is the \ac{UVDAR} system \cite{walter_icra2020, uvdd1, uvdd2}, which detects blinking \ac{UV} \acp{LED} placed on the target in camera image frames.
The active blinking of the \acp{LED} allows our solution to be used even in poor light~conditions, such as darkness, where the AprilTag detector fails to provide sufficient measurements.
The \ac{UV} \acp{LED} placed on the \ac{USV} board are shown in \reffig{fig:sensors}.
The \ac{UVDAR} system provides measurements of the position and orientation of the target.

\subsection{Trajectory Planner}
\label{sec:trajectory_planner}
The trajectory planner used for experimental verification is based on linear \ac{MPC}, with detailed descriptions provided in \cite{trajectory_planner}.
This planner utilizes estimated and predicted \ac{USV} states (\refsec{sec:state_estimator_predictor}) to align the \ac{UAV} trajectory with the motion of the \ac{USV}.
When the task is to follow the \ac{USV}, the \ac{UAV} maintains a desired distance above the \ac{USV} board and promptly responds to changes in \ac{USV} movement, such as those induced by waves.
This responsiveness is important in applications where the \ac{UAV} is tethered to the \ac{USV} using a power supply cable to recharge \ac{UAV} batteries \cite{talke2018CatenaryTetherShape}.
Failure of the \ac{UAV} to react to the motion of the \ac{USV} on waves poses a risk of the cable pulling the \ac{UAV} and destabilizing it. 

The landing task includes even more challenges.
Firstly, the \ac{UAV} must track the \ac{USV} steadily.
Subsequently, the \ac{UAV} begins descending towards the \ac{USV} board.
In the final phase, the \ac{UAV} lands on the \ac{USV} board at a predefined vertical velocity relative to the \ac{USV} velocity.
This controlled descent is crucial for a safe landing, as it ensures that the vertical and touchdown velocity of the \ac{UAV} are regulated throughout the entire maneuver, regardless of the vertical motion of the \ac{USV} in the waves.
Maintaining a static descending velocity despite the vertical motion of the \ac{USV} carries a significant risk, as waves may push the \ac{USV} towards the \ac{UAV}, which significantly increases the \ac{UAV} touchdown velocity, potentially causing damage to the \ac{UAV}.

\section{\uppercase{Verification}}
The proposed approach was first verified in simulations and subsequently deployed in real-world experiments. 
Moreover, we compare our results with those obtained using the state-of-the-art method \cite{polvara_6dof}.
The simulations were performed in a realistic robotic simulator Gazebo, extended by the \ac{VRX} simulator \cite{bingham19toward}, which provides a realistic simulation of the harsh marine environment with large waves, as the \ac{UAV}-\ac{USV} collaboration in such conditions is our main motivation.
A video attachment supporting the results of this paper is available at \url{https://mrs.fel.cvut.cz/papers/towards-uav-usv-collaboration}.
The novelty of our approach lies in \ac{USV} state estimation and prediction using our proposed novel \ac{USV} model containing wave dynamics (\refsec{sec:mathematical_model}).
Therefore, we evaluate our estimation and prediction using \ac{RMSE}.
The \reftab{tab:rmse_estimated} presents the \ac{RMSE} of the estimated \ac{USV} states using our approach computed from the performed simulations, in which the \ac{UAV} follows and lands on the \ac{USV} in harsh environment.
The \ac{RMSE} values are computed with respect to the sensors used for the estimation.

\begin{table}[!b]
  \caption{RMSE of the estimated USV states using the approach proposed in this paper with respect to the individual sensors.}
  \small
  \centering
    \begin{tabular}{lcccc}
    \hline
    sensor & \makecell{RMSE\\ $(x,y,z)$\\m} & \makecell{RMSE\\ $(\phi,\theta,\psi)$\\rad} & \makecell{RMSE\\ $(u,v,w)$\\m/s} & \makecell{RMSE\\ $(p,q,r)$\\rad/s}\\
    \hline
    GPS & 0.989 & - & 0.985 & -\\
    IMU & - & 0.011 & - & 0.536\\
    UVDAR & 0.425 & 0.124 & 0.977 & 1.034\\
    AprilTag & 0.088 & 0.052 & 0.848 & 0.606\\
    \hline
    \end{tabular}
  \label{tab:rmse_estimated}
\end{table}

\begin{table}[!b]
    \caption{RMSE of the estimated USV states using the approach proposed in this paper compared to a state-of-the-art method.}
    \small
  \centering
    \begin{tabular}{lcccc}
    \hline
    method & \makecell{RMSE\\ $(x,y,z)$\\m} & \makecell{RMSE\\ $(\phi,\theta,\psi)$\\rad} & \makecell{RMSE\\ $(u,v,w)$\\m/s} & \makecell{RMSE\\ $(p,q,r)$\\rad/s}\\ 
    \hline
    \sota{} & 0.313 & 0.056 & 0.580 & 0.329\\
    our approach & 0.116 & 0.017
  & 0.303 & 0.235\\
    \hline
    \end{tabular}
  \label{tab:rmse_sota_comparison}
\end{table}

The estimation of position $(x, y, z)$ and corresponding linear velocities $(u, v, w)$ using \ac{GPS} yields results with the largest \ac{RMSE}.
A better estimation for the states $(x, y, z)$ and $(u, v, w)$ compared to \ac{GPS} is achieved through the \ac{UVDAR} system.
The \ac{RMSE} for states $(x, y, z)$ is half the size when using the \ac{UVDAR} system compared to the \ac{GPS}, and for states $(u, v, w)$, the \ac{RMSE} using the \ac{UVDAR} system is slightly smaller than when using the \ac{GPS} sensor.
However, for the states $(\phi, \theta, \psi)$ and corresponding angular velocities $(p, q, r)$, the \ac{UVDAR} yields the largest \ac{RMSE} among all sensors.
The most accurate estimation of the states $(\phi, \theta, \psi)$ is attained using \ac{IMU} data, as indicated by its minimal \ac{RMSE}.
The smallest \ac{RMSE} for $(x, y, z)$ and $(u, v, w)$ is achieved using the AprilTag detector.
Moreover, the estimation of states $(\phi, \theta, \psi)$ and $(p, q, r)$ using the AprilTag detector results in a \ac{RMSE} that is half as large as that of the \ac{UVDAR} system.
However, the \ac{RMSE} of the AprilTag detector for states $(\phi, \theta, \psi)$ remains four and a half times larger than that of the \ac{IMU} for the same states.

\begin{figure}[!tb]
    \centering
    \includegraphics[width=\linewidth]{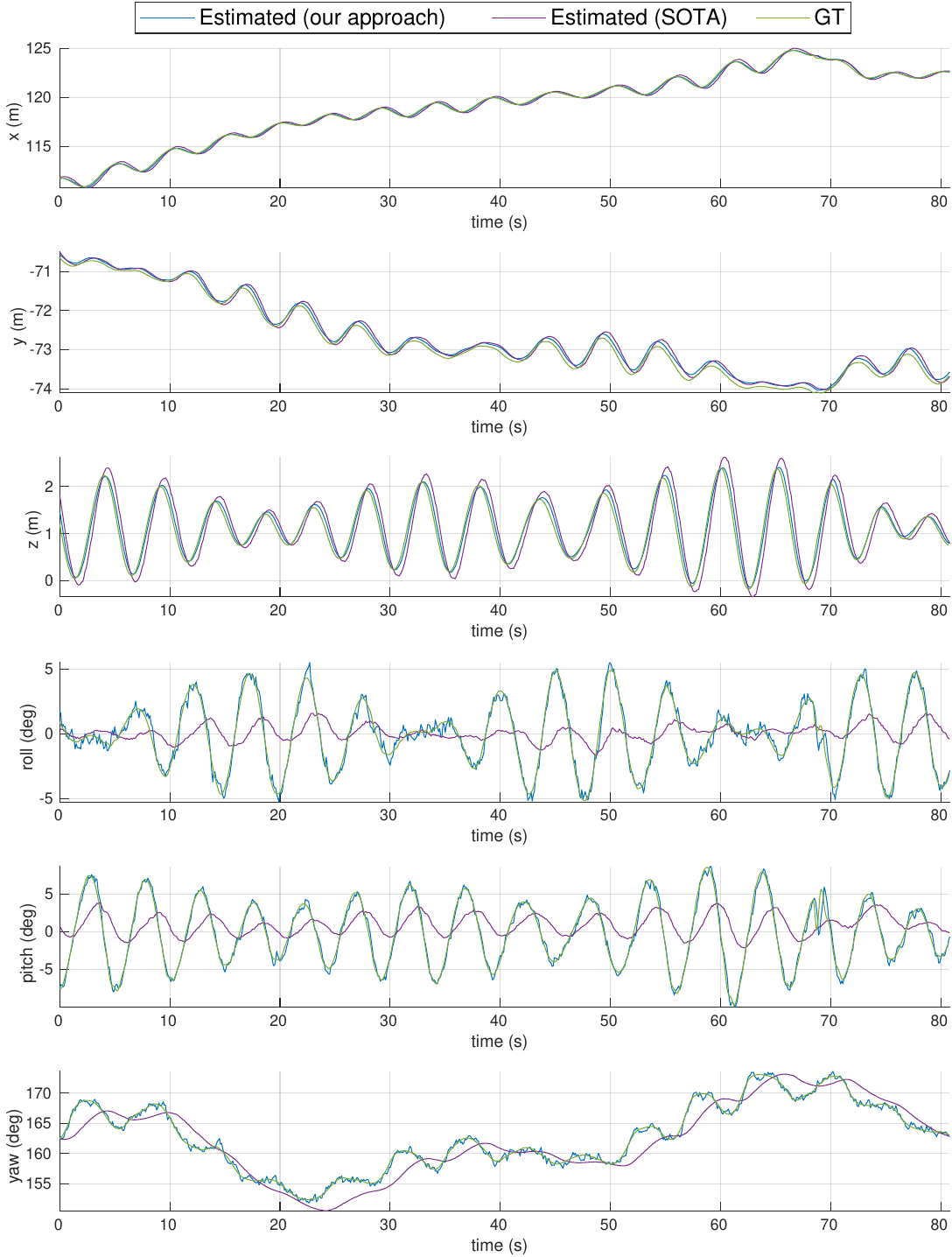}
    \caption{Estimated USV position $\vec{p}=(x, y, z)$ and orientation $\vec{\Theta}=(\phi, \theta, \psi)$ using the method proposed in this paper.}
    \label{fig:estimations}
\end{figure}

The \reffig{fig:estimations} shows the estimated \ac{USV} states from one of the many performed simulations, in which our approach fused data from all sensors.
The graphs illustrate that all estimated states correspond to the \ac{GT} values.
The \ac{RMSE} of estimated states using our approach is provided in the \reftab{tab:rmse_sota_comparison}.
We compared our approach with the most relevant state-of-the-art method \cite{polvara_6dof}, which we call \sota{} in this paper, as depicted in \reffig{fig:estimations}.
Our approach has smaller \ac{RMSE} for all \ac{USV} states (\reftab{tab:rmse_sota_comparison}).
The main difference is observed in the estimation of orientation $(\phi,\theta,\psi)$, where our approach achieves more than three times smaller \ac{RMSE} compared to the \sota{}.
This substantial improvement in orientation estimation can be attributed to the overall \ac{UAV}-\ac{USV} collaboration system by the incorporation waves in the mathematical model and the fusion of data from multiple sensors, features lacking in \sota{}.
The impact of this difference is evident in \reffig{fig:estimations} in the estimation of heave $z$, roll $\phi$, and pitch $\theta$, where our approach exhibits closer alignment with \ac{GT} values compared to \sota{}.

\begin{figure}[!tb]
    \centering
    \includegraphics[width=\linewidth]{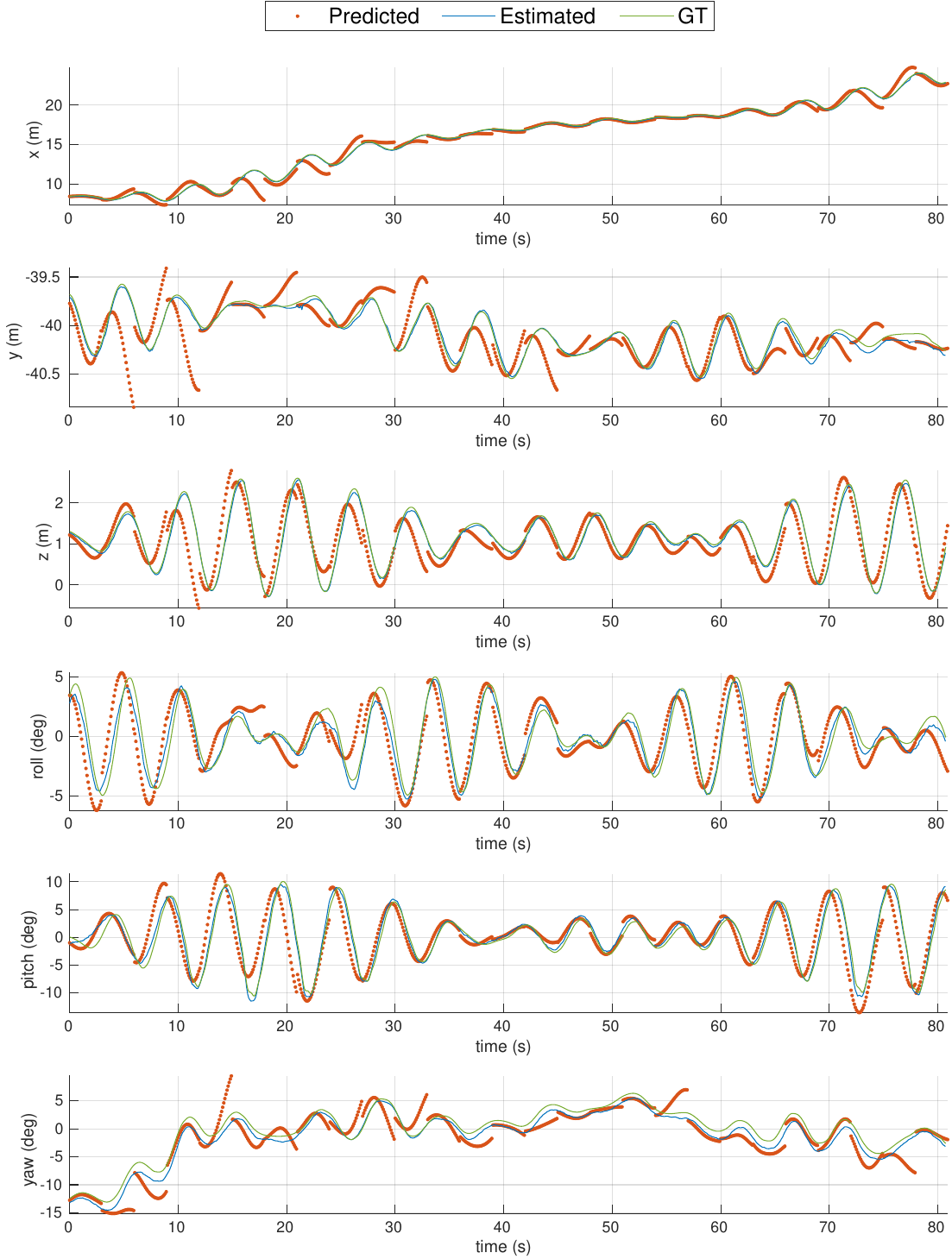}
    \caption{Predicted USV position $\vec{p}=(x, y, z)$ and orientation $\vec{\Theta}=(\phi, \theta, \psi)$ using the method proposed in this paper.}
    \label{fig:predicted_usv_states_pos_rot}
\end{figure}

The predicted \ac{USV} states $(x,y,z,\phi,\theta,\psi)$ from one of the performed simulations are shown in \reffig{fig:predicted_usv_states_pos_rot}.
The two-second predictions are computed every two seconds.
The figure illustrates that the predictions initially deviate more from the \ac{GT} values at the beginning of the simulation.
However, as the estimation progresses over time, the predicted \ac{USV} states become increasingly accurate.
The \ac{RMSE} of predictions for states $(x, y, z)$ proposed in \reftab{tab:rmse_predicted} corresponds to the \ac{RMSE} of estimation of these states using only the GPS sensor (\reftab{tab:rmse_estimated}).
The \ac{RMSE} of predictions for states $(\phi,\theta,\psi)$ corresponds to the \ac{RMSE} of these states estimated using \ac{UVDAR} system (\reftab{tab:rmse_estimated}). 
These results demonstrate the applicability of computed predictions to \ac{UAV} trajectory planner.

\begin{table}[!tb]
  \caption{RMSE of predicted and estimated USV states using the method proposed in this paper.}
  \small
  \centering
    \begin{tabular}{lcc}
    \hline
    USV states (our approach) & \makecell{RMSE\\ $(x,y,z)$\\m} & \makecell{RMSE\\ $(\phi,\theta,\psi)$\\rad}\\
    \hline
    predicted states & 0.737 & 0.196\\
    estimated states & 0.116 & 0.017\\
    \hline
    \vspace{0.001em}
    \end{tabular}
  \label{tab:rmse_predicted}
\end{table}

\subsection{Real-World Experiments}
\begin{figure}[!b]
    \centering
    \input{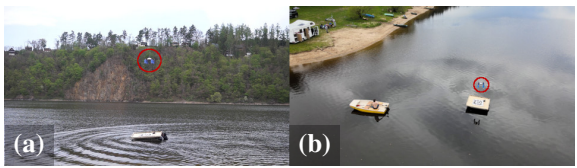}
    \caption{The figure depicts snapshots from the performed real-world experiments, in which the UAV (marked with a red circle) followed the USV (a) and landed on the USV (b).}
    \label{fig:snapshots}
\end{figure}

To analyze the real-world performance and to show robustness to real uncertainties and disturbances, we deployed the presented approach in real-world experiments.
During the first set of real-world experiments, the overall system demonstrated robust performance, allowing the \ac{UAV} to repeatedly follow the \ac{USV} and land on the \ac{USV}'s deck.
In these experiments, the \ac{USV} was manually forced to perform wave motions, as depicted in \reffig{fig:snapshots} (a).
The \ac{UAV} was equipped with an onboard computer, \ac{GPS} sensor, and cameras for the AprilTag detector and the \ac{UVDAR} system (see \cite{HertJINTHW_paper, MRS2022ICUAS_HW} for details).
The \ac{USV}'s deck, featuring the AprilTag and \ac{UV} \acp{LED}, is detailed in \reffig{fig:sensors}.
Additionally, the \ac{USV} carried the \ac{MRS} boat unit containing \ac{IMU} and \ac{GPS} sensors.
The detailed description of the used \ac{UAV} and \ac{USV} sensors is provided in \refsec{sec:usv_onboard_sensors} and \ref{sec:uav_onboard_sensors}.

\begin{figure}[!tb]
    \centering
    \input{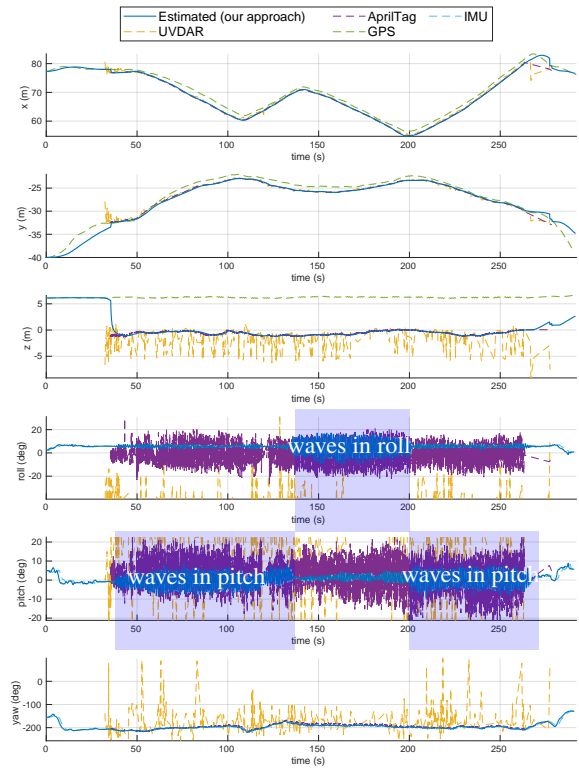}
    \caption{Estimated USV states using our approach in one of the performed real-world experiments.}
    \label{fig:estimated_usv_states_rw}
\end{figure}

The estimated \ac{USV} states from one of the performed real-world experiments are depicted in \reffig{fig:estimated_usv_states_rw}.
Initially, the \ac{UAV} approached the \ac{USV} while \ac{USV} states were estimated using received \ac{GPS} and \ac{IMU} data.
Subsequently, the \ac{UAV} onboard sensors improved the estimation, as evident in the graphs of \ac{USV} position $(x,y,z)$ (\reffig{fig:estimated_usv_states_rw}).
Then, the \ac{UAV} followed the \ac{USV}, which performed artificially induced wave motion that can be noticed especially in roll and pitch graphs.
The pitch was affected by waves from 37~s to 130~s and from 200~s to 264~s, while the roll was affected from 130~s to 200~s.

The \ac{UAV} was also able to successfully land on the \ac{USV}, as depicted in \reffig{fig:snapshots} (b).
The \ac{USV} was towed by another boat, while the \ac{UAV} followed it using a trajectory planner (\refsec{sec:trajectory_planner}) that utilized estimated and predicted \ac{USV} states.
When the conditions for a safe landing were met, the \ac{UAV} initiated a landing maneuver, which was successfully completed within 6~s.

\section{\uppercase{Conclusion}}
In this paper, we introduce a novel onboard \ac{UAV} approach designed to facilitate tight collaboration between \ac{UAV} and \ac{USV} in harsh marine environments, such as following or landing maneuvers.
The main contribution of the proposed solution is the \ac{USV} state estimator and predictor, operating in 6 \acp{DOF}, which uses our novel mathematical \ac{USV} model incorporating wave dynamics.
The state estimator fuses data from multiple \ac{UAV} and \ac{USV} sensors, ensuring accurate estimation across various real-world conditions.
Subsequently, the estimated \ac{USV} states are fed into the state predictor, which utilizes the mathematical \ac{USV} model to predict future \ac{USV} states in 6 \acp{DOF}. 
We verified the overall system through extensive simulations and compared the results of the proposed approach with the state-of-the-art method.
The proposed approach was also deployed in real-world experiments, where the \ac{UAV} was able to repeatedly follow the \ac{USV} and land on the \ac{USV}'s deck.
Future work will focus on developing a nonlinear mathematical model of the \ac{USV} to better capture its dynamics on a wavy water surface. 
Moreover, the proposed approach is planned to be integrated into an autonomous \ac{UAV}-\ac{USV} team designed for garbage removal and water quality monitoring.

\section*{\uppercase{Acknowledgements}}
This work was funded by the Czech Science Foundation (GA\v{C}R) under research project no. 23-07517S, by the European Union under the project Robotics and advanced industrial production (reg. no. CZ.02.01.01/00/22\_008/0004590), and by CTU grant no SGS23/177/OHK3/3T/13.

\bibliographystyle{apalike}
{\small
\bibliography{main}}

\begin{thebibliography}{}

\bibitem[Abujoub et~al., 2018]{Abujoub2018_landing}
Abujoub, S., McPhee, J., Westin, C., and Irani, R.~A. (2018).
\newblock Unmanned aerial vehicle landing on maritime vessels using signal prediction of the ship motion.
\newblock In {\em OCEANS 2018 MTS/IEEE Charleston}, pages 1--9.

\bibitem[Aissi et~al., 2020]{usv_uav_docking}
Aissi, M., Moumen, Y., Berrich, J., Bouchentouf, T., Bourhaleb, M., and Rahmoun, M. (2020).
\newblock Autonomous solar usv with an automated launch and recovery system for uav: State of the art and design.
\newblock In {\em 2020 IEEE 2nd International Conference on Electronics, Control, Optimization and Computer Science (ICECOCS)}, pages 1--6.

\bibitem[Aniceto et~al., 2018]{marine_mammals_monitoring}
Aniceto, A.~S., Biuw, M., Lindstrøm, U., Solbø, S.~A., Broms, F., and Carroll, J. (2018).
\newblock Monitoring marine mammals using unmanned aerial vehicles: quantifying detection certainty.
\newblock {\em Ecosphere}, 9(3):e02122.

\bibitem[Baca et~al., 2021]{baca2021mrs}
Baca, T., Petrlik, M., Vrba, M., Spurny, V., Penicka, R., Hert, D., and Saska, M. (2021).
\newblock {The MRS UAV System: Pushing the Frontiers of Reproducible Research, Real-world Deployment, and Education with Autonomous Unmanned Aerial Vehicles}.
\newblock {\em Journal of Intelligent {\&} Robotic Systems}, 102(26):1--28.

\bibitem[Bingham et~al., 2019]{bingham19toward}
Bingham, B., Aguero, C., McCarrin, M., Klamo, J., Malia, J., Allen, K., Lum, T., Rawson, M., and Waqar, R. (2019).
\newblock Toward maritime robotic simulation in {Gazebo}.
\newblock In {\em Proceedings of MTS/IEEE OCEANS Conference}, Seattle, WA.

\bibitem[Fossen, 2011]{Fossen2011}
Fossen, T.~I. (2011).
\newblock {\em Handbook of Marine Craft Hydrodynamics and Motion Control}.
\newblock John Wiley \& Sons, United Kingdom, first edition edition.

\bibitem[Gupta et~al., 2023]{uav_usv_landing_parakh}
Gupta, P.~M., Pairet, E., Nascimento, T., and Saska, M. (2023).
\newblock Landing a uav in harsh winds and turbulent open waters.
\newblock {\em IEEE Robotics and Automation Letters}, 8(2):744--751.

\bibitem[Han and Ma, 2021]{water_pollution}
Han, Y. and Ma, W. (2021).
\newblock Automatic monitoring of water pollution based on the combination of uav and usv.
\newblock In {\em 2021 IEEE 4th International Conference on Electronic Information and Communication Technology (ICEICT)}, pages 420--424.

\bibitem[{Hert} et~al., 2023]{HertJINTHW_paper}
{Hert}, D., {Baca}, T., {Petracek}, P., {Kratky}, V., {Penicka}, R., {Spurny}, V., {Petrlik}, M., {Vrba}, M., {Zaitlik}, D., {Stoudek}, P., {Walter}, V., {Stepan}, P., {Horyna}, J., {Pritzl}, V., {Sramek}, M., {Ahmad}, A., {Silano}, G., {Bonilla Licea}, D., {Stibinger}, P., {Nascimento}, T., and {Saska}, M. (2023).
\newblock {MRS Drone: A Modular Platform for Real-World Deployment of Aerial Multi-Robot Systems}.
\newblock {\em Journal of Intelligent {\&} Robotic Systems}.

\bibitem[{Hert} et~al., 2022]{MRS2022ICUAS_HW}
{Hert}, D., {Baca}, T., {Petracek}, P., {Kratky}, V., {Spurny}, V., {Petrlik}, M., {Vrba}, M., {Zaitlik}, D., {Stoudek}, P., {Walter}, V., {Stepan}, P., {Horyna}, J., {Pritzl}, V., {Silano}, G., {Bonilla Licea}, D., {Stibinger}, P., {Penicka}, R., {Nascimento}, T., and {Saska}, M. (2022).
\newblock {MRS Modular UAV Hardware Platforms for Supporting Research in Real-World Outdoor and Indoor Environments}.
\newblock In {\em 2022 International Conference on Unmanned Aircraft Systems (ICUAS)}, pages 1264--1273. IEEE.

\bibitem[Kalman, 1960]{kalman1960new}
Kalman, R.~E. (1960).
\newblock A new approach to linear filtering and prediction problems.
\newblock {\em Journal of Basic Engineering}, 82(1):35--45.

\bibitem[Keller and Ben-Moshe, 2022]{keller2022}
Keller, A. and Ben-Moshe, B. (2022).
\newblock A robust and accurate landing methodology for drones on moving targets.
\newblock {\em Drones}, 6(4).

\bibitem[Krogius et~al., 2019]{krogius2019iros}
Krogius, M., Haggenmiller, A., and Olson, E. (2019).
\newblock Flexible layouts for fiducial tags.
\newblock In {\em {IEEE/RSJ} International Conference on Intelligent Robots and Systems {(IROS)}}.

\bibitem[Lee et~al., 2019]{lee2019}
Lee, S., Lee, J., Lee, S., Choi, H., Kim, Y., Kim, S., and Suk, J. (2019).
\newblock Sliding mode guidance and control for uav carrier landing.
\newblock {\em IEEE Transactions on Aerospace and Electronic Systems}, 55(2):951--966.

\bibitem[Marconi et~al., 2002]{MARCONI200221}
Marconi, L., Isidori, A., and Serrani, A. (2002).
\newblock Autonomous vertical landing on an oscillating platform: an internal-model based approach.
\newblock {\em Automatica}, 38(1):21--32.

\bibitem[Meng et~al., 2019]{MENG2019474}
Meng, Y., Wang, W., Han, H., and Ban, J. (2019).
\newblock A visual/inertial integrated landing guidance method for uav landing on the ship.
\newblock {\em Aerospace Science and Technology}, 85:474--480.

\bibitem[Murphy et~al., 2006]{usv_uav_hurricane_wilma2}
Murphy, R., Stover, S., Pratt, K., and Griffin, C. (2006).
\newblock Cooperative damage inspection with unmanned surface vehicle and micro unmanned aerial vehicle at {Hurricane Wilma}.
\newblock In {\em 2006 IEEE/RSJ International Conference on Intelligent Robots and Systems}, pages 9--9.

\bibitem[Murphy et~al., 2008]{usv_uav_hurricane_wilma}
Murphy, R.~R., Steimle, E., Griffin, C., Cullins, C., Hall, M., and Pratt, K. (2008).
\newblock Cooperative use of unmanned sea surface and micro aerial vehicles at {Hurricane Wilma}.
\newblock {\em Journal of Field Robotics}, 25(3):164--180.

\bibitem[Olson, 2011]{olson2011tags}
Olson, E. (2011).
\newblock {AprilTag}: A robust and flexible visual fiducial system.
\newblock In {\em {IEEE} International Conference on Robotics and Automation ({ICRA})}, pages 3400--3407. IEEE.

\bibitem[Polvara et~al., 2018]{polvara_6dof}
Polvara, R., Sharma, S., Wan, J., Manning, A., and Sutton, R. (2018).
\newblock Vision-based autonomous landing of a quadrotor on the perturbed deck of an unmanned surface vehicle.
\newblock {\em Drones}, 2(2).

\bibitem[Prochazka, 2023]{trajectory_planner}
Prochazka, O. (2023).
\newblock Trajectory planning for autonomous landing of a multirotor helicopter on a boat.
\newblock Master's thesis, Faculty of Electrical Engineering, Czech Technical University in Prague.

\bibitem[Román et~al., 2023]{water_quality_monitoring}
Román, A., Tovar-Sánchez, A., Gauci, A., Deidun, A., Caballero, I., Colica, E., D’Amico, S., and Navarro, G. (2023).
\newblock Water-quality monitoring with a uav-mounted multispectral camera in coastal waters.
\newblock {\em Remote Sensing}, 15(1).

\bibitem[Talke et~al., 2018]{talke2018CatenaryTetherShape}
Talke, K.~A., De~Oliveira, M., and Bewley, T. (2018).
\newblock Catenary tether shape analysis for a uav - usv team.
\newblock In {\em 2018 IEEE/RSJ International Conference on Intelligent Robots and Systems (IROS)}, pages 7803--7809.

\bibitem[Tran and Ahn, 2019]{TRAN201967}
Tran, Q.~V. and Ahn, H.-S. (2019).
\newblock Multi-agent localization of a common reference coordinate frame: An extrinsic approach.
\newblock {\em IFAC-PapersOnLine}, 52(20):67--72.
\newblock 8th IFAC Workshop on Distributed Estimation and Control in Networked Systems NECSYS 2019.

\bibitem[Venugopalan et~al., 2012]{Venugopalan2012}
Venugopalan, T.~K., Taher, T., and Barbastathis, G. (2012).
\newblock Autonomous landing of an unmanned aerial vehicle on an autonomous marine vehicle.
\newblock In {\em 2012 Oceans}, pages 1--9.

\bibitem[Walter et~al., 2018a]{uvdd2}
Walter, V., N.Staub, Saska, M., and Franchi, A. (2018a).
\newblock Mutual localization of {UAVs} based on blinking ultraviolet markers and {3D} time-position {Hough} transform.
\newblock In {\em 14th IEEE International Conference on Automation Science and Engineering (CASE 2018)}.

\bibitem[Walter et~al., 2018b]{uvdd1}
Walter, V., Saska, M., and Franchi, A. (2018b).
\newblock Fast mutual relative localization of {UAVs} using ultraviolet {LED} markers.
\newblock In {\em 2018 International Conference on Unmanned Aircraft System (ICUAS 2018)}.

\bibitem[{Walter} et~al., 2020]{walter_icra2020}
{Walter}, V., {Vrba}, M., and {Saska}, M. (2020).
\newblock On training datasets for machine learning-based visual relative localization of micro-scale {UAVs}.
\newblock In {\em 2020 IEEE International Conference on Robotics and Automation (ICRA)}, pages 10674--10680.

\bibitem[Wang and Olson, 2016]{wang2016iros}
Wang, J. and Olson, E. (2016).
\newblock {AprilTag} 2: Efficient and robust fiducial detection.
\newblock In {\em {IEEE/RSJ} International Conference on Intelligent Robots and Systems {(IROS)}}.

\bibitem[Xu et~al., 2020]{Xu2020_vision}
Xu, Z.-C., Hu, B.-B., Liu, B., Wang, X., and Zhang, H.-T. (2020).
\newblock Vision-based autonomous landing of unmanned aerial vehicle on a motional unmanned surface vessel.
\newblock In {\em 2020 39th Chinese Control Conference (CCC)}, pages 6845--6850.

\bibitem[Yang et~al., 2021]{yang2021}
Yang, L., Liu, Z., Wang, X., Wang, G., Hu, X., and Xi, Y. (2021).
\newblock Autonomous landing of a rotor unmanned aerial vehicle on a boat using image-based visual servoing.
\newblock In {\em 2021 IEEE International Conference on Robotics and Biomimetics (ROBIO)}, pages 1848--1854.

\bibitem[Zhang et~al., 2021]{zhang2021}
Zhang, H.-T., Hu, B.-B., Xu, Z., Cai, Z., Liu, B., Wang, X., Geng, T., Zhong, S., and Zhao, J. (2021).
\newblock Visual navigation and landing control of an unmanned aerial vehicle on a moving autonomous surface vehicle via adaptive learning.
\newblock {\em IEEE Transactions on Neural Networks and Learning Systems}, 32(12):5345--5355.

\end{thebibliography}

\begin{acronym}
  \acro{CNN}[CNN]{Convolutional Neural Network}
  \acro{IR}[IR]{infrared}
  \acro{GNSS}[GNSS]{Global Navigation Satellite System}
  \acro{UT}[UT]{Unscented Transform}
  \acro{API}[API]{Application Programming Interface}
  \acro{CTU}[CTU]{Czech Technical University}
  \acroplural{DOF}[DOFs]{Degrees of Freedom}
  \acro{DOF}[DOF]{Degree of Freedom}
  \acro{FOV}[FOV]{Field of View}
  \acro{GPS}[GPS]{Global Positioning System}
  \acro{IMU}[IMU]{Inertial Measurement Unit}
  \acro{LKF}[LKF]{Linear Kalman Filter}
  \acro{LTI}[LTI]{Linear time-invariant}
  \acro{LiDAR}[LiDAR]{Light Detection and Ranging}
  \acro{MPC}[MPC]{Model Predictive Control}
  \acro{MRS}[MRS]{Multi-robot Systems}
  \acro{ROS}[ROS]{Robot Operating System}
  \acro{SLAM}[SLAM]{Simultaneous Localization And Mapping}
  \acro{UAV}[UAV]{Unmanned Aerial Vehicle}
  \acro{UGV}[UGV]{Unmanned Ground Vehicle}
  \acro{UKF}[UKF]{Unscented Kalman Filter}
  \acro{USV}[USV]{Unmanned Surface Vehicle}
  \acro{RMSE}[RMSE]{Root Mean Square Error}
  \acro{UVDAR}[UVDAR]{UltraViolet Direction And Ranging}
  \acro{UV}[UV]{UltraViolet}
  \acro{LED}[LED]{Light-Emitting Diode}
  \acro{VRX}[VRX]{Virtual RobotX}
  \acro{WAM-V}[WAM-V]{Wave Adaptive Modular Vessel}
  \acro{GT}[GT]{Ground Truth}
  \acro{UTM}[UTM]{Universal Transverse Mercator}
  \acro{VTOL}[VTOL]{Vertical Take-Off and Landing}
  \acro{FFT}[FFT]{Fast Fourier Transform}
\end{acronym}

\end{document}